\documentclass{article}

\usepackage{arxiv}

\usepackage[utf8]{inputenc} 
\usepackage[T1]{fontenc}    
\usepackage{hyperref}       
\usepackage{url}            
\usepackage{booktabs}       
\usepackage{amsfonts}       
\usepackage{nicefrac}       
\usepackage{microtype}      
\usepackage{lipsum}		
\usepackage{graphicx}
\usepackage{natbib}
\usepackage{doi}
\usepackage{graphicx}
\usepackage{xcolor}
\usepackage{algorithm}
\usepackage{algpseudocode}
\usepackage{tikz,amsmath} 
\usepackage{subcaption}
\usepackage{standalone}
\usetikzlibrary{positioning,fit,calc,backgrounds}
\usepackage{caption}

\pgfdeclarelayer{background}
\pgfsetlayers{background,main}
\title{Controlled Territory and Conflict Tracking (CONTACT): (Geo-)Mapping Occupied Territory from Open Source Intelligence}

\author{ \href{https://orcid.org/0000-0002-4966-0494}{\includegraphics[scale=0.06]{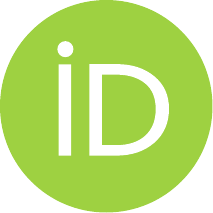}\hspace{1mm}Paul K. Mandal $^{1,2,*}$} \\
	\texttt{mandal@utexas.edu} \\
    \And
    \href{https://orcid.org/0009-0008-6875-6281}{\includegraphics[scale=0.06]{orcid.pdf}\hspace{1mm}Cole Leo $^{3}$} \\
    \texttt{cleo@t2s-solutions.com}
	\And
    Connor Hurley $^{3}$\\
	\texttt{churley@t2s-solutions.com} \\
    \And
    \\
    $^1$ Neurint LLC, Middletown DE, USA \\
    $^2$ The University of Texas at Austin, Austin TX, USA \\
    $^3$ T2S Solutions, Aberdeen MD, USA \\
    $^*$ Corresponding Author \\
}

\renewcommand{\headeright}{}
\renewcommand{\undertitle}{}

\hypersetup{
pdftitle={Horizontal Federated Computer Vision},
pdfsubject={cs.LG, cs.AI},
pdfauthor={Paul K. Mandal, Cole Leo, and Connor Hurley},
pdfkeywords={},
}

\begin{document}
\maketitle

\begin{abstract}
Open-source intelligence provides a stream of unstructured textual data that can inform assessments of territorial control. We present CONTACT, a framework for territorial control prediction using large language models (LLMs) and minimal supervision. We evaluate two approaches: SetFit, an embedding-based few-shot classifier, and a prompt tuning method applied to BLOOMZ-560m, a multilingual generative LLM. Our model is trained on a small hand-labeled dataset of news articles covering ISIS activity in Syria and Iraq, using prompt-conditioned extraction of control-relevant signals such as military operations, casualties, and location references. We show that the BLOOMZ-based model outperforms the SetFit baseline, and that prompt-based supervision improves generalization in low-resource settings. CONTACT demonstrates that LLMs fine-tuned using few-shot methods can reduce annotation burdens and support structured inference from open-ended OSINT streams. Our code is available at \hyperlink{https://github.com/PaulKMandal/CONTACT/}{https://github.com/PaulKMandal/CONTACT/}.
\end{abstract}

\section{Introduction}
\label{sec:intro}

Identifying patterns of territorial control in conflict zones remains a core challenge for military intelligence, humanitarian response, and peacekeeping operations. Traditional methods rely on manual analysis of open-source reports, resulting in significant delays and incomplete situational awareness—especially in dynamic and contested environments. This limitation has become more critical in modern asymmetric conflicts, where control is fluid, decentralized, and communicated through fragmented or ambiguous open-source signals.

Despite the expanding availability of open source intelligence (OSINT), real-time territorial inference remains a difficult challenge. A recent survey of terrorism-related OSINT extraction outlines the technical bottlenecks in this space, identifying data acquisition, enrichment, and inference as distinct but under-integrated phases of current systems, many of which lack domain adaptability and temporal sensitivity \citep{chaudhary2022open}. Additionally, geopolitical ambiguity and linguistic variation further complicate the automated interpretation of territorial control. Large language models (LLMs), when applied in these contexts, have been shown to exhibit geopolitical bias—returning inconsistent or conflicting representations of territorial boundaries depending on the language or phrasing of the query \citep{li2024land}. These inconsistencies can undermine the reliability of inference in multilingual environments, where sovereignty and legitimacy are highly contested. Recent work by \cite{castillo2023role} further highlights the risk that LLMs may reinforce implicit territorial claims embedded in their training data, thereby shaping or legitimizing sovereignty through model outputs. These findings underscore the sociopolitical risks of applying LLMs in conflict monitoring without domain-specific grounding and control.

Our paper makes the following contributions: (1) a lightweight, open-source scraping utility for collecting conflict-related news articles from archived web sources; (2) a hand-labeled dataset of ISIS-related news reports annotated with VIINA-style territorial control indicators; (3) a comparison of few-shot classification using SetFit versus prompt-tuned BLOOMZ with parameter-efficient fine-tuning; (4) evidence that embedding label definitions into prompts improves performance in low-resource, multi-label settings; and (5) training code for CONTACT using both SetFit and BLOOMZ.

\section{Prior Work}

Efforts to infer territorial control from structured data have traditionally relied on statistical modeling frameworks. One influential approach by \cite{anders2020territorial} conceptualizes territorial control as a latent variable and employs Hidden Markov Models (HMMs) \citep{baum1966statistical,rabiner1989tutorial} to estimate shifts in control based on observable features such as patterns in rebel tactics and spatial-temporal dynamics of conflict events. These models provide a principled way to incorporate uncertainty and structure over time, but they assume the availability of clean, event-coded input streams. In operational environments where analysts rely on raw or unstructured text, these assumptions often fail. As a result, there is a growing recognition that purely statistical models require upstream support from natural language processing systems capable of extracting relevant signals from open text sources.

A major step toward automating territorial inference from unstructured text came with the development of VIINA (Violent Incident Information from News Articles), introduced by \cite{zhukov2023near}. The original system employed a long short-term memory (LSTM) architecture \citep{hochreiter1997long} to classify and extract violent event information from news sources during the Russian invasion of Ukraine. VIINA 2.0 switched to a bidirectional encoder representations from transformers (BERT) architecture \citep{devlin2019bert}, enabling more accurate extraction of event attributes such as actor type, location, and casualty information \citep{zhukov2023viina}. The extracted location spans are then geocoded using the Yandex Geocoding API, allowing the system to map events to precise geographic coordinates. VIINA is trained on both Ukrainian- and Russian-language news sources, allowing it to cover narratives from both sides of the conflict. 

However, VIINA is highly specialized for the conflict in Ukraine. It is trained exclusively on Ukrainian and Russian news sources and depends on conflict-specific labels and hand-curated features. While its BERT-based model performs well in extracting structured information, it does not support zero-shot or few-shot generalization. This means the model must be retrained or heavily fine-tuned to adapt to new conflicts or shifts in labeling standards. Unlike generative models that support in-context learning, BERT cannot dynamically adjust to new prompts or classification tasks without annotated training data. As a result, VIINA’s utility is constrained when applied to new domains or multilingual reporting environments, where rapid adaptation and minimal supervision are essential.

More recent efforts explore the use of generative language models. SmartBook, introduced by \cite{reddy2023smartbook}, frames situation report generation as a question answering (QA) task. It uses a RoBERTa-large encoder \citep{liu2019roberta} trained on SQuAD 2.0 \citep{rajpurkar-etal-2018-know} and Natural Questions \citep{kwiatkowski-etal-2019-natural} to extract answers from a news corpus in response to predefined strategic questions. Answer spans are mapped back to their source sentences, which are treated as extracted claims. To reduce false positives, a secondary RoBERTa-large classifier—trained on Natural Questions and WikiQA \citep{yang-etal-2015-wikiqa}—is used to validate each answer-context pair. The top five validated contexts are then passed to a summarization model for report generation. While SmartBook demonstrates the potential of LLMs in intelligence workflows, its emphasis is on high-level synthesis rather than the extraction of operational indicators such as territorial presence, actor affiliation, or contested locations. Its outputs are not structured for integration into downstream models that rely on event-level tagging, geolocation, or time-series analysis. As such, SmartBook is well-suited for summarizing knowns but less effective at surfacing unknowns—especially when those unknowns require fine-grained, structured inference.

Other recent work explores adjacent uses of language models in military and geographic domains. COA-GPT applies reinforcement learning to fine-tune generative models for course-of-action development in military planning tasks \citep{goecks2024coa}. Separately, GPT-4 has been shown to possess coarse but nontrivial representations of global geography \citep{roberts2023gpt4geo}. However, neither system addresses the structured extraction of control-relevant indicators from multilingual OSINT streams. This leaves a gap between flexible language modeling capabilities and operationally grounded territorial inference—one that CONTACT seeks to address.

\section{Dataset}
To train and evaluate the performance of CONTACT, we constructed a hand-curated dataset of open-source news articles covering the expansion of ISIS in Syria and Iraq. The articles span the period from 2015-2019, during which ISIS gained and lost control over key territories including Mosul, Raqqa, and Deir ez-Zor. This period is well-suited for territorial inference tasks due to frequent changes in control and the availability of reporting on military operations, governance, and violence against civilians.

Our dataset consists of 20 news articles, manually selected to ensure coverage of diverse events and locations. To test the model’s few-shot generalization capabilities, we use 15 articles for training and 5 for evaluation. Each article was annotated at the sentence level using a simplified version of the VIINA tagging scheme. We used the 5 labels shown in table \ref{tab:viina-labels}.

\begin{table}[h]
\centering
\begin{tabular}{ll}
\toprule
\textbf{Label} & \textbf{Description} \\
\midrule
\texttt{t\_mil} & Event is about war/military operations \\
\texttt{t\_loc} & Event report includes reference to specific location \\
\texttt{t\_milcas} & Event report mentions military casualties \\
\texttt{t\_civcas} & Event report mentions civilian casualties \\
\texttt{t\_isis\_vic} & ISIS won the conflict \\
\bottomrule
\end{tabular}
\vspace{0.5em}
\caption{VIINA-style labels used for annotating conflict-relevant information.}
\label{tab:viina-labels}
\end{table}

All labels were manually annotated to ensure consistency. Articles were drawn from publicly available English-language news sources and cover a variety of conflict-relevant topics, including frontline engagements, airstrikes, and territorial takeovers.

\section{Methods}

\subsection{Data Pipeline}

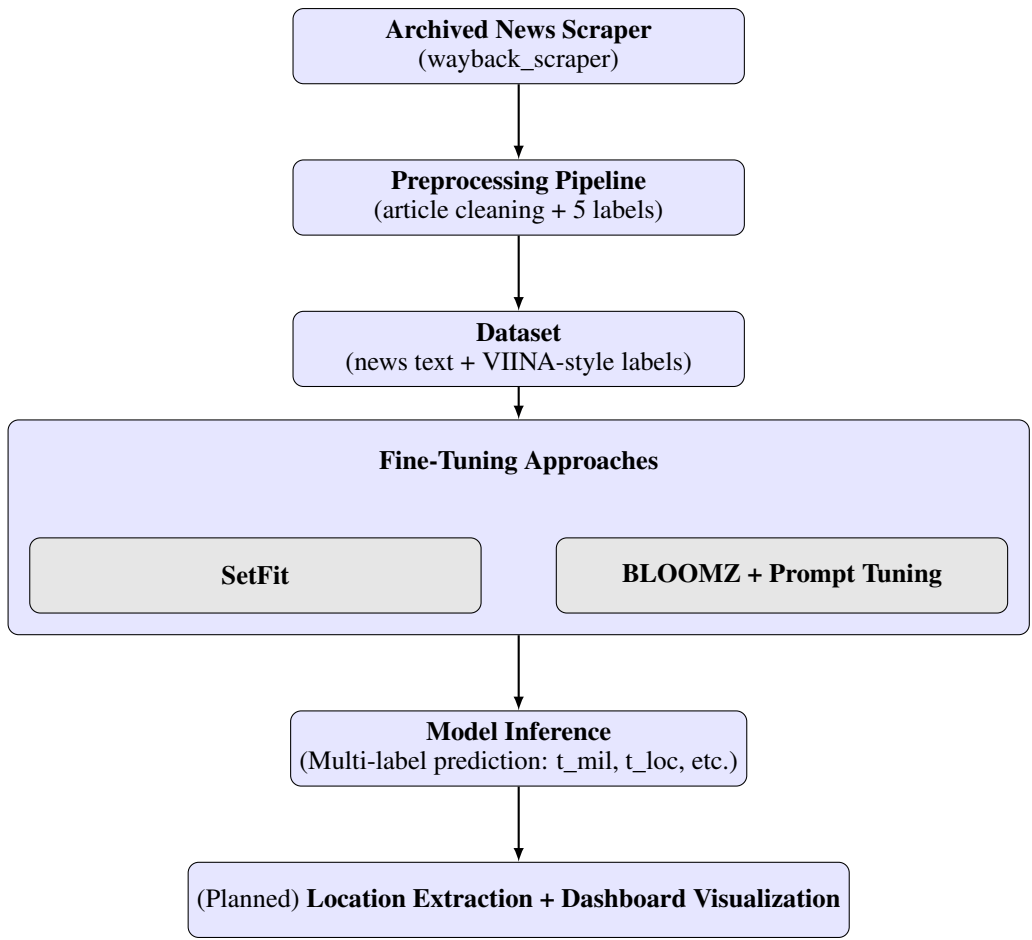
\begin{figure}[htbp]
  \centering
\begin{tikzpicture}[
  block/.style={
    draw=black,
    fill=blue!10,
    rounded corners,
    align=center,
    minimum width=6cm,
    minimum height=1cm
  },
    block2/.style={
    draw=black,
    fill=gray!20,
    rounded corners,
    align=center,
    minimum width=6cm,
    minimum height=1cm
  },
  ftbox/.style={
    draw=black,
    fill=blue!10,
    rounded corners,
    inner sep=8pt
  },
  arrow/.style={-latex, thick}
]
  \node[block] (scraper)
        {\textbf{Archived News Scraper}\\(wayback\_scraper)};
  \node[block, below=1cm of scraper] (preproc)
        {\textbf{Preprocessing Pipeline}\\(article cleaning + 5 labels)};
  \node[block, below=1cm of preproc] (dataset)
        {\textbf{Dataset}\\(news text + VIINA-style labels)};

  \node[block2, below=2cm of dataset, xshift=-3.5cm] (setfit)
        {\textbf{SetFit}};
  \node[block2, below=2cm of dataset, xshift= 3.5cm] (bloomz)
        {\textbf{BLOOMZ + Prompt Tuning}};

  \node[align=center] (ftitle)
       at ($(setfit.north)!0.5!(bloomz.north)+(0,1cm)$)
       {\textbf{Fine‑Tuning Approaches}};

  \begin{scope}[on background layer]
    \node[ftbox, fit=(ftitle)(setfit)(bloomz)] (finebox) {};
  \end{scope}

  \node[block, below=1cm of finebox] (inference)
        {\textbf{Model Inference}\\(Multi‑label prediction: t\_mil, t\_loc, etc.)};
  \node[block, below=1cm of inference] (future)
        {(Planned) \textbf{Location Extraction + Dashboard Visualization}};

  \draw[arrow] (scraper)   -- (preproc);
  \draw[arrow] (preproc)   -- (dataset);
  \draw[arrow] (dataset)   -- (finebox);
  \draw[arrow] (finebox)   -- (inference);
  \draw[arrow] (inference) -- (future);
\end{tikzpicture}
  \caption{Pipeline for the CONTACT framework: articles are scraped from archived news sources, preprocessed, and labeled with territorial control indicators. The resulting dataset is used to fine-tune two models: a SetFit classifier based on sentence embeddings, and a BLOOMZ model using prompt tuning. Fine-tuned models perform multi-label inference, with future extensions including location extraction and dashboard integration.}
  \label{fig:contact-pipeline}
\end{figure}

Because open-source articles are frequently removed, geo-restricted, or structurally altered over time, we developed a scraping framework that combines live retrieval with archival recovery. Articles were first fetched using a wrapper built on newspaper3k, which downloads and parses full text from HTML content \citep{newspaper3k}. For sources that were inaccessible or modified, we used a custom archival tool based on the Wayback Machine \citep{wayback_machine}. This tool leverages the waybackpy API \citep{Mahanty_waybackpy_2022} to retrieve time-specific snapshots of web pages by URL and date. Our full processing pipeline is illustrated in Figure~\ref{fig:contact-pipeline}.

To handle retrieval of archived content, we implemented a custom tool called \texttt{wayback\_scraper}, which wraps the waybackpy API and integrates with the newspaper3k library for article extraction. This utility retrieves date-specific snapshots of news pages from the Wayback Machine, parses them into structured Article objects, and supports keyword filtering over both titles and body text. It accepts configurable request timeouts and user-agent headers to improve scraping resilience. We release wayback\_scraper as an open-source utility to support dataset construction from archived open-source news content.\footnote{\url{https://github.com/PaulKMandal/CONTACT/blob/main/scraper/scrape_utils.py}}

Following retrieval, all articles are normalized by truncating the main body text to a maximum of 512 characters to meet downstream model input limits. Empty or malformed entries are dropped. The result is a minimally preprocessed, text-rich dataset suitable for prompt-based training or supervised extraction pipelines.

\subsection{Few-Shot Classification with SetFit}

For our first approach, we fine-tuned a sentence-transformer model using the SetFit framework \citep{tunstall2022efficient} for multi-label classification. We initialized the model with \texttt{paraphrase-mpnet-base-v2}, a pretrained encoder from the SentenceTransformers library \citep{reimers-2019-sentence-bert}, and applied a one-vs-rest logistic classification head for each of the five VIINA-style labels.

Each article in the training set was encoded into a dense vector using the frozen transformer backbone. The resulting embeddings were passed to a classification head trained using cosine similarity loss. The dataset was mapped into multi-label binary vectors, and label encoding followed the same ordering as in Table~\ref{tab:viina-labels}. We trained using 14 articles and evaluated on 5, with batch size 1 and 20 training iterations per epoch.

\subsection{Prompt-Tuned BLOOMZ with Prompt Tuning}

Our second approach uses BLOOMZ 560M \citep{muennighoff2023crosslingual}, a causal language model which we fine-tuned using prompt tuning \citep{lester-etal-2021-power}. We treat territorial inference as a sequence-to-sequence task in which the model receives a truncated news article and is prompted to generate the relevant control labels.

We initialize prompt tuning with 8 virtual tokens using the \texttt{PromptTuningConfig} class from the Parameter Efficient Fine-Tuning library from huggingface \citep{peft}. The prompt is initialized with the following descriptive string, which is prepended to every training instance:

\begin{quote}
\texttt{Give each of these following labels a 0 if false and a 1 if true:} \\
\texttt{t\_mil - Event is about war/military operations,} \\
\texttt{t\_loc - Event report includes reference to specific location,} \\
\texttt{t\_milcas - Event report mentions military casualties,} \\
\texttt{t\_civcas - Event report mentions civilian casualties,} \\
\texttt{t\_isis\_vic - ISIS won the conflict}
\end{quote}

Each training input is constructed by concatenating this prompt with the article text using the format:

\begin{quote}
\texttt{text : [ARTICLE TEXT] Label :}
\end{quote}

The expected output is a comma-separated string of label names (e.g., \texttt{t\_mil, t\_loc, t\_isis\_vic}). Input-label pairs are tokenized, concatenated, and masked such that loss is computed only over the label portion. Inputs are padded on the left to accommodate BLOOMZ’s decoder-only architecture. We used a maximum input length of 256 tokens, batch size of 1, learning rate of 3e-2, and trained for one epoch using AdamW with linear learning rate warmup. The base model weights remain frozen during training. This setup allows BLOOMZ to learn label-conditioned generation using only a small set of examples while avoiding full model fine-tuning.

\section{Results}

We evaluated both models using per-label binary accuracy on a held-out test set of five manually labeled articles. Each article was annotated with five VIINA-style labels, and each label was treated as a separate binary classification task. Model predictions were compared against ground truth annotations, and performance was averaged across all labels and all examples.

The \textbf{SetFit} model performed poorly, achieving an average per-label accuracy of 40\%. Upon inspection, we found that it only predicted the two most frequent labels on the test set—\texttt{t\_mil} and \texttt{t\_loc}—for every article, regardless of content. It failed to predict any of the less common labels. This failure mode suggests that the model collapsed to a trivial baseline conditioned only on label frequency, which is a known issue in few-shot multi-label classification settings when using pooled sentence embeddings and cosine similarity-based objectives. The inability to distinguish among label types likely stems from a lack of context-specific representation and insufficient signal in the small training set to separate semantically overlapping events.

In contrast, the \textbf{BLOOMZ + Prompt Tuning} model achieved 100\% accuracy across all five labels and all test articles. The prompt-tuned decoder-only architecture effectively translated open-domain text into the target label space, even when the relevant information was implicit or distributed across multiple sentences. By embedding label definitions directly in the prompt, the model was able to contextualize its generation and disambiguate subtle cues associated with territorial control, military action, and outcome-specific reporting. Unlike SetFit, BLOOMZ made varied predictions that reflected the actual structure of the inputs.

Although the BLOOMZ results are encouraging, they must be interpreted cautiously. The dataset is small, and the test set includes only five examples, which may not capture the diversity of real-world reporting. Furthermore, the prompt-based formulation introduces dependencies on phrasing and input formatting that could affect generalization. Nonetheless, the experiment provides initial evidence that parameter-efficient fine-tuning of instruction-following models offers a promising approach for low-resource, OSINT-based territorial inference.

\section{Discussion}

Our experiments demonstrate that prompt-conditioned language models provide a viable path toward predicting territorial control from OSINT when labeled data is limited. A key insight from our work is the importance of embedding label definitions and semantic context directly into the model input. The BLOOMZ model, for instance, consistently outperformed the embedding-based SetFit approach, largely due to its access to contextual prompts that defined the structure and meaning of each label. This highlights the practical value of exposing the model to clear label definitions in the prompt during fine-tuning.

Shortly after this study, we developed a related system, CONTACT-SALUTE, trained on 100 synthetic spot reports from the Ukraine conflict. These reports were structured using the SALUTE format—Size, Activity, Location, Unit, Time, and Equipment—which is commonly used in military intelligence reporting. CONTACT-SALUTE also included a secondary language model dedicated to extracting the location of the incident from the text. While the system and dataset are proprietary and cannot be released, the approach demonstrated that standardized, structured reporting can improve model performance. Future versions of CONTACT should include location extraction as a core capability and support visual analytics through an integrated dashboard and actively track controlled territory.

Reducing the annotation burden remains a major barrier to scaling territorial inference systems. Emerging techniques in \textit{deep active learning} offer potential solutions by optimizing which samples to label next based on model uncertainty or informativeness \citep{croicu2024deep}. However, these techniques still rely on the presence of human annotators to verify uncertain predictions or correct model outputs, which may not be feasible at scale in fast-moving conflict environments.

Another area for future improvement lies in the foundation models themselves. Much of our fine-tuning and evaluation work was conducted in late 2023, prior to the emergence of more capable LLMs that have since demonstrated stronger zero-shot and few-shot performance across multilingual and geopolitically sensitive tasks. Applying these newer models, especially those tuned for factuality, could further boost accuracy and and improve generalization across conflict domains.

Ultimately, while our results are preliminary, we show that the annotation burden for designing OSINT-based territorial monitoring systems can be significantly reduced using few-shot learning. This enables faster adaptation to new conflict settings without the need for extensive manual labeling, making such systems more practical to develop and deploy.

\bibliographystyle{apalike}
\bibliography{egbib}

\begin{thebibliography}{}

\bibitem[Anders, 2020]{anders2020territorial}
Anders, T. (2020).
\newblock Territorial control in civil wars: Theory and measurement using machine learning.
\newblock {\em Journal of Peace Research}, 57(6):701--714.

\bibitem[Baum and Petrie, 1966]{baum1966statistical}
Baum, L.~E. and Petrie, T. (1966).
\newblock Statistical inference for probabilistic functions of finite state markov chains.
\newblock {\em The annals of mathematical statistics}, 37(6):1554--1563.

\bibitem[Castillo-Eslava et~al., 2023]{castillo2023role}
Castillo-Eslava, F., Mougan, C., Romero-Reche, A., and Staab, S. (2023).
\newblock The role of large language models in the recognition of territorial sovereignty: An analysis of the construction of legitimacy.
\newblock {\em arXiv preprint arXiv:2304.06030}.

\bibitem[Chaudhary and Bansal, 2022]{chaudhary2022open}
Chaudhary, M. and Bansal, D. (2022).
\newblock Open source intelligence extraction for terrorism-related information: A review.
\newblock {\em Wiley Interdisciplinary Reviews: Data Mining and Knowledge Discovery}, 12(5):e1473.

\bibitem[Croicu, 2024]{croicu2024deep}
Croicu, M. (2024).
\newblock Deep active learning for data mining from conflict text corpora.
\newblock {\em arXiv preprint arXiv:2402.01577}.

\bibitem[Devlin et~al., 2019]{devlin2019bert}
Devlin, J., Chang, M.-W., Lee, K., and Toutanova, K. (2019).
\newblock Bert: Pre-training of deep bidirectional transformers for language understanding.
\newblock In {\em Proceedings of the 2019 conference of the North American chapter of the association for computational linguistics: human language technologies, volume 1 (long and short papers)}, pages 4171--4186.

\bibitem[Goecks and Waytowich, 2024]{goecks2024coa}
Goecks, V.~G. and Waytowich, N. (2024).
\newblock {COA-GPT}: Generative pre-trained transformers for accelerated course of action development in military operations.
\newblock In {\em 2024 International Conference on Military Communication and Information Systems (ICMCIS)}, pages 01--10. IEEE.

\bibitem[Hochreiter and Schmidhuber, 1997]{hochreiter1997long}
Hochreiter, S. and Schmidhuber, J. (1997).
\newblock Long short-term memory.
\newblock {\em Neural computation}, 9(8):1735--1780.

\bibitem[{Internet Archive}, 2025]{wayback_machine}
{Internet Archive} (2025).
\newblock Wayback machine.
\newblock \url{https://web.archive.org/}.
\newblock Accessed: 2025-04-09.

\bibitem[Kwiatkowski et~al., 2019]{kwiatkowski-etal-2019-natural}
Kwiatkowski, T., Palomaki, J., Redfield, O., Collins, M., Parikh, A., Alberti, C., Epstein, D., Polosukhin, I., Devlin, J., Lee, K., Toutanova, K., Jones, L., Kelcey, M., Chang, M.-W., Dai, A.~M., Uszkoreit, J., Le, Q., and Petrov, S. (2019).
\newblock Natural questions: A benchmark for question answering research.
\newblock {\em Transactions of the Association for Computational Linguistics}, 7:452--466.

\bibitem[Lester et~al., 2021]{lester-etal-2021-power}
Lester, B., Al-Rfou, R., and Constant, N. (2021).
\newblock The power of scale for parameter-efficient prompt tuning.
\newblock In Moens, M.-F., Huang, X., Specia, L., and Yih, S. W.-t., editors, {\em Proceedings of the 2021 Conference on Empirical Methods in Natural Language Processing}, pages 3045--3059, Online and Punta Cana, Dominican Republic. Association for Computational Linguistics.

\bibitem[Li et~al., 2024]{li2024land}
Li, B., Haider, S., and Callison-Burch, C. (2024).
\newblock This land is \{Your, My\} land: Evaluating geopolitical biases in language models through territorial disputes.
\newblock {\em 2024 Annual Conference of the North American Chapter of the Association for Computational Linguistics (NAACL)}.

\bibitem[Liu et~al., 2019]{liu2019roberta}
Liu, Y., Ott, M., Goyal, N., Du, J., Joshi, M., Chen, D., Levy, O., Lewis, M., Zettlemoyer, L., and Stoyanov, V. (2019).
\newblock Roberta: A robustly optimized bert pretraining approach.
\newblock {\em arXiv preprint arXiv:1907.11692}.

\bibitem[Mahanty, 2022]{Mahanty_waybackpy_2022}
Mahanty, A. (2022).
\newblock {waybackpy}.
\newblock Accessed: 2025-04-09.

\bibitem[Mangrulkar et~al., 2022]{peft}
Mangrulkar, S., Gugger, S., Debut, L., Belkada, Y., Paul, S., and Bossan, B. (2022).
\newblock Peft: State-of-the-art parameter-efficient fine-tuning methods.
\newblock \url{https://github.com/huggingface/peft}.

\bibitem[Muennighoff et~al., 2023]{muennighoff2023crosslingual}
Muennighoff, N., Wang, T., Sutawika, L., Roberts, A., Biderman, S., Le~Scao, T., Bari, M.~S., Shen, S., Yong, Z.~X., Schoelkopf, H., et~al. (2023).
\newblock Crosslingual generalization through multitask finetuning.
\newblock In {\em Proceedings of the 61st Annual Meeting of the Association for Computational Linguistics (Volume 1: Long Papers)}, pages 15991--16111.

\bibitem[Ou-Yang, 2016]{newspaper3k}
Ou-Yang, L. (2016).
\newblock newspaper3k: Article scraping and curation.
\newblock \url{https://newspaper.readthedocs.io/}.
\newblock Accessed: 2025-04-09.

\bibitem[Rabiner, 1989]{rabiner1989tutorial}
Rabiner, L.~R. (1989).
\newblock A tutorial on hidden markov models and selected applications in speech recognition.
\newblock {\em Proceedings of the IEEE}, 77(2):257--286.

\bibitem[Rajpurkar et~al., 2018]{rajpurkar-etal-2018-know}
Rajpurkar, P., Jia, R., and Liang, P. (2018).
\newblock Know what you don`t know: Unanswerable questions for {SQ}u{AD}.
\newblock In {\em Proceedings of the 56th Annual Meeting of the Association for Computational Linguistics (Volume 2: Short Papers)}, pages 784--789, Melbourne, Australia. Association for Computational Linguistics.

\bibitem[Reddy et~al., 2023]{reddy2023smartbook}
Reddy, R.~G., Lee, D., Fung, Y.~R., Nguyen, K.~D., Zeng, Q., Li, M., Wang, Z., Voss, C., and Ji, H. (2023).
\newblock Smartbook: {AI}-assisted situation report generation for intelligence analysts.
\newblock {\em arXiv preprint arXiv:2303.14337}.

\bibitem[Reimers and Gurevych, 2019]{reimers-2019-sentence-bert}
Reimers, N. and Gurevych, I. (2019).
\newblock Sentence-bert: Sentence embeddings using siamese bert-networks.
\newblock In {\em Proceedings of the 2019 Conference on Empirical Methods in Natural Language Processing}. Association for Computational Linguistics.

\bibitem[Roberts et~al., 2023]{roberts2023gpt4geo}
Roberts, J., L{\"u}ddecke, T., Das, S., Han, K., and Albanie, S. (2023).
\newblock {GPT4GEO}: How a language model sees the world's geography.
\newblock {\em arXiv preprint arXiv:2306.00020}.

\bibitem[Tunstall et~al., 2022]{tunstall2022efficient}
Tunstall, L., Reimers, N., Jo, U. E.~S., Bates, L., Korat, D., Wasserblat, M., and Pereg, O. (2022).
\newblock Efficient few-shot learning without prompts.
\newblock {\em arXiv preprint arXiv:2209.11055}.

\bibitem[Yang et~al., 2015]{yang-etal-2015-wikiqa}
Yang, Y., Yih, W.-t., and Meek, C. (2015).
\newblock {W}iki{QA}: A challenge dataset for open-domain question answering.
\newblock In M{\`a}rquez, L., Callison-Burch, C., and Su, J., editors, {\em Proceedings of the 2015 Conference on Empirical Methods in Natural Language Processing}, pages 2013--2018, Lisbon, Portugal. Association for Computational Linguistics.

\bibitem[Zhukov and Ayers, 2023]{zhukov2023viina}
Zhukov, Y. and Ayers, N. (2023).
\newblock {VIINA 2.0: Violent Incident Information from News Articles on the 2022 Russian Invasion of Ukraine}.
\newblock \url{https://github.com/zhukovyuri/VIINA}.
\newblock Cambridge, MA: Harvard University. Accessed April 8, 2025.

\bibitem[Zhukov, 2023]{zhukov2023near}
Zhukov, Y.~M. (2023).
\newblock Near-real time analysis of war and economic activity during russia’s invasion of ukraine.
\newblock {\em Journal of Comparative Economics}, 51(4):1232--1243.

\end{thebibliography}

\end{document}